\documentclass[sigconf, nonacm]{acmart}

\usepackage{amsmath}
\usepackage{graphicx}
\usepackage{booktabs}
\usepackage{pifont}
\usepackage{enumerate}

\usepackage{multirow,float}
\usepackage{bm}   
\usepackage[linesnumbered,ruled,vlined]{algorithm2e}
\usepackage{xcolor}

\SetCommentSty{mycommfont}
\usepackage{tikz,pgfplots}
\usetikzlibrary{pgfplots.groupplots}
\usetikzlibrary{patterns}
\usetikzlibrary{matrix}
\definecolor{reb}{RGB}{0,0,0}

\usetikzlibrary{external}
\newcommand\vldbdoi{XX.XX/XXX.XX}
\newcommand\vldbpages{XXX-XXX}
\newcommand\vldbvolume{14}
\newcommand\vldbissue{1}
\newcommand\vldbyear{2020}
\newcommand\vldbauthors{\authors}
\newcommand\vldbtitle{\shorttitle} 
\newcommand\vldbavailabilityurl{https://github.com/amazon-research/tgl}
\newcommand\vldbpagestyle{plain}

\newcommand{\cmark}{\ding{51}}

\begin{document}
\title{TGL: A General Framework for Temporal GNN Training on Billion-Scale Graphs}

\author{Hongkuan Zhou}
\authornote{The work was performed during an internship at AWS AI Lab.}
\affiliation{%
  \institution{University of Southern California}
}
\email{hongkuaz@usc.edu}

\author{Da Zheng}
\affiliation{%
  \institution{AWS AI}
}
\email{dzzhen@amazon.com}

\author{Israt Nisa}
\affiliation{%
  \institution{AWS AI}
}
\email{nisisrat@amazon.com}

\author{Vasileios Ioannidis}
\affiliation{%
  \institution{AWS AI}
}
\email{ivasilei@amazon.com}

\author{Xiang Song}
\affiliation{%
  \institution{AWS AI}
}
\email{xiangsx@amazon.com}

\author{George Karypis}
\affiliation{%
  \institution{AWS AI}
}
\email{gkarypis@amazon.com}

\begin{abstract}
Many real world graphs contain time domain information. 
Temporal Graph Neural Networks capture temporal information as well as structural and contextual information in the generated dynamic node embeddings. Researchers have shown that these embeddings achieve state-of-the-art performance in many different tasks. In this work, we propose TGL, a unified framework \textcolor{reb}{for large-scale offline Temporal Graph Neural Network training} where users can compose various Temporal Graph Neural Networks with simple configuration files. TGL comprises five main components, a temporal sampler, a mailbox, a node memory module, a memory updater, and a message passing engine. \textcolor{reb}{We design a Temporal-CSR data structure and a parallel sampler to efficiently sample temporal neighbors to form training mini-batches.} We propose a novel random chunk scheduling technique that mitigates the problem of obsolete node memory \textcolor{reb}{when training with a large batch size}. \textcolor{reb}{To address the limitations of current TGNNs only being evaluated on small-scale datasets, we introduce two large-scale real-world datasets with 0.2 and 1.3 billion temporal edges}. We evaluate the performance of TGL on \textcolor{reb}{four small-scale datasets with a single GPU} and the two large datasets with multiple GPUs for both link prediction and node classification tasks. We compare TGL with the open-sourced code of five methods and show that TGL achieves similar or better accuracy with an average of $13\times$ speedup. Our temporal parallel sampler achieves an average of $173\times$ speedup 
on a multi-core CPU compared with the baselines. On a 4-GPU machine, TGL can train one epoch of more than one billion temporal edges within \textcolor{reb}{1-10 hours}. To the best of our knowledge, this is the first work \textcolor{reb}{that proposes a general framework for large-scale Temporal Graph Neural Networks training on multiple GPUs}.
\end{abstract}
\maketitle

\pagestyle{\vldbpagestyle}
\begingroup\small\noindent\raggedright\textbf{PVLDB Reference Format:}\\
\vldbauthors. \vldbtitle. PVLDB, \vldbvolume(\vldbissue): \vldbpages, \vldbyear.\\
\href{https://doi.org/\vldbdoi}{doi:\vldbdoi}
\endgroup
\begingroup
\renewcommand\thefootnote{}\footnote{\noindent
This work is licensed under the Creative Commons BY-NC-ND 4.0 International License. Visit \url{https://creativecommons.org/licenses/by-nc-nd/4.0/} to view a copy of this license. For any use beyond those covered by this license, obtain permission by emailing \href{mailto:info@vldb.org}{info@vldb.org}. Copyright is held by the owner/author(s). Publication rights licensed to the VLDB Endowment. \\
\raggedright Proceedings of the VLDB Endowment, Vol. \vldbvolume, No. \vldbissue\ %
ISSN 2150-8097. \\
\href{https://doi.org/\vldbdoi}{doi:\vldbdoi} \\
}\addtocounter{footnote}{-1}\endgroup

\ifdefempty{\vldbavailabilityurl}{}{
\vspace{.3cm}
\begingroup\small\noindent\raggedright\textbf{PVLDB Artifact Availability:}\\
The source code, data, and/or other artifacts have been made available at \url{\vldbavailabilityurl}.
\endgroup
}

\section{Introduction}
\label{sec:intro}

Graph Neural Networks (GNNs) have proven to be powerful and reliable method in representation learning on static graphs and are widely used in many academic and industrial problems. There exist well-developed libraries like DGL \cite{wang2019dgl} and PyG \cite{pyg} that allow users to quickly and efficiently implement GNN variants \textcolor{reb}{for static graphs} and deploy them to CPUs, GPUs, or even distributed systems. There are also \textcolor{reb}{multiple} benchmark dataset collections like OGB \cite{hu2020ogb,hu2021ogblsc} that provide large-scale and wide-ranging datasets to evaluate the performance of GNN variants \textcolor{reb}{for static graphs}.

However, many real-world graphs are dynamic. For example, \textcolor{reb}{in a social network new users join over time and users interact with each other on posts and send messages}. 
\textcolor{reb}{In a knowledge graph, new events appear and are only valid for specific periods of time}.
\textcolor{reb}{The dynamics in the user-item graph reveal important information in identifying abusive behaviors \cite{wang2021bipartite}.}
To capture the evolving nature on dynamic graphs, recently, researchers \cite{tgat,tgn,evolvegcn,dysat} have developed Temporal Graph Neural Networks (TGNNs) which jointly learn the temporal, structural, and contextual relationships on \textcolor{reb}{dynamic} graphs. 
Like static GNNs, TGNNs encode graph information at a given time into dynamic node embeddings.
With the additional temporal information, TGNNs outperform static GNNs on link prediction tasks, dynamic node classification tasks, and many other tasks on dynamic graphs.
\textcolor{reb}{These works on temporal graph representation learning are developed using different frameworks with different levels of optimizations and parallelization
and are evaluated on small dynamic graphs which only contain less than ten thousand nodes and one million edges.}

\begin{figure}[b!]
    \centering
    \input{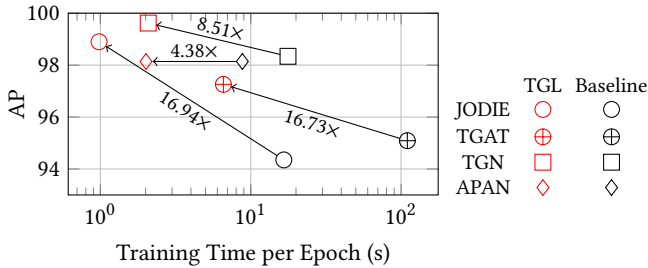}
    \caption{Accuracy and per epoch training time of TGL compared with the baselines on the Wikipedia dataset (600 batch size).}
    \label{fig:aptime}
\end{figure}
\setlength{\marginparwidth}{15mm}

\textcolor{reb}{Many real-world dynamic graphs, such as social network graphs and knowledge graphs, usually have millions of nodes and billions of edges.} It is challenging to scale \textcolor{reb}{TGNN} training to large graphs for multiple reasons. First, the additional temporal dependency requires training to be done chronologically. TGNNs also use more expensive neighbor samplers which select temporal neighbors based on the timestamps of the interactions between nodes. Moreover, different TGNNs capture the temporal information using different strategies like node memory and snapshot.
\textcolor{reb}{Existing graph deep learning frameworks like DGL and PyG do not provide efficient data structure, sampler, and message passing primitive for dynamic graphs, which requires users to implement extra modules to compose TGNN models. In addition, it is also challenging to design an efficient and versatile framework that is capable of unifying the different schemes of different TGNN variants.}
\textcolor{reb}{Recently, PyTorch Geometric Temporal (PyGT) \cite{pygt} attempted to design a library for dynamic and temporal geometric deep learning on top of PyG.} However, PyGT only supports discrete time snapshot-based method and full batch training on small-scale spatial-temporal graphs. 

To fill these gaps, we develop TGL, the first general framework for \textcolor{reb}{large-scale offline TGNNs training. In this work,} we focus on the widely used edge-timestamped dynamic graphs where each edge is associated with a timestamp. TGL supports all TGNN variants that aggregate and refine information from maintained states or features of selected temporal neighbors. 
\textcolor{reb}{The survey \cite{kazemi2020representation} categorizes dynamic graphs into Continuous Time Dynamic Graphs (CTDGs) and Discrete Time Dynamic Graphs (DTDGs) based on the continuous or discrete quantity of the timestamps. However, we believe that DTDGs are essentially CTDGs with granulated timestamps. Hence, we design TGL to support \textcolor{reb}{the more general} CTDGs and evaluate TGL by comparing the performance of TGNN variants targeting both CTDGs and DTDGs in the experiments.} Our main contributions are
\begin{itemize}
    \item We design a unified framework that supports efficient training on most TGNN architectures by studying the characteristic of a diverse set of TGNNs variants including snapshot-based TGNNs \cite{dysat,evolvegcn}, time encoding-based TGNNs \cite{tgat,tgn,apan}, and memory-based TGNNs \cite{jodie,tgn,apan,dyrep}.
    \item We design a CSR-based data structure for rapid access to temporal neighbors and a parallel sampler that support different temporal neighbor sampling algorithms. Our parallel sampler can quickly locate the temporal edges to sample from by maintaining auxiliary pointer arrays. 
    \item We propose a novel random chunk scheduling technique that overcomes the deprivation of intra-dependency when training with a large batch size for the methods using node memory, which enables multi-GPU training on large-scale dynamic graphs.
    \item \textcolor{reb}{To better compare the performance of various TGNN methods, we introduce} two large-scale datasets \textcolor{reb}{with billions of edges -- the GDELT and MAG datasets} which represent dynamic graphs with long time duration and dynamic graphs with larger number of nodes.
    \item We compare the performance of TGL with the baseline open-sourced codes on \textcolor{reb}{two small-scale datasets}. TGL achieves similar or higher accuracy for all baseline methods with an average speedup of $13\times$ as shown in Figure \ref{fig:aptime}. On the large-scale datasets, TGL achieves an average of $2.3\times$ speedup when using 4 GPUs.
\end{itemize}
\section{Temporal Graph Neural Networks}

\begin{table}[t]
    \centering
    \setlength{\tabcolsep}{1.5mm}
    \begin{tabular}{r|c|c|c|c|c|c|c|c}
        & \cite{evolvegcn} & \cite{dysat} & \cite{jodie} & \cite{dyrep} & \cite{dysat} & \cite{tgat} & \cite{tgn} & \cite{apan}\\
        \toprule
        snapshot & \cmark & \cmark & & & \cmark & & &\\
        time encoding & & & & & & \cmark & \cmark & \cmark\\
        memory & & & \cmark & \cmark & & & \cmark & \cmark
    \end{tabular}
    \caption{Strategies used by various TGNN variants.}
    \label{tab:tgnnsum}
\end{table}

\begin{table}[b]
    \centering
    \setlength{\tabcolsep}{1mm}
    \begin{tabular}{c|l}
        Symbol & Description\\
        \toprule
        $u,v,i,j$ & Nodes in dynamic graphs\\
        $e$ & Edges in dynamic graphs\\
        $\bm v_v,\bm e_{ij}$ & Node feature of $v$ and edge feature of edge $ij$\\
        $\mathcal{N}(v)$ & Set of past neighbors of node $v$\\
        $m_e^{uv},m_e^{vu}$ & Mails generated at the source and destination nodes\\
        $\bm s_v$ & Node memory of node $v$\\
        $t^-_v$ & Time when $\bm s_v$ is updated\\
        $\Phi(\cdot)$ & Time encoder\\
    \end{tabular}
    \caption{Notation used in this paper}
    \label{tab:notation}
\end{table}

TGNNs generate dynamic node embeddings by adding components like time encoder and node memory in the message passing flow \textcolor{reb}{or combining information from multiple consecutive graph snapshots.}. In general, TGNNs generate dynamic node embeddings by iteratively processing the information gathered from temporal neighbors, similarly to static GNNs.
To capture the additional temporal dependencies, TGNNs usually process the neighbor information by three methods: 1) group the neighbors in the past according to their time and learn sequences from the groups (snapshot-based TGNNs), 2) add additional time encoding to each past neighbor, and 3) maintain node memory which summarizes the current state of each node (memory-based TGNNs). 
Table \ref{tab:tgnnsum} shows different strategies used by different TGNN variants. Note that some TGNNs \cite{tgn,apan} use combinations of multiple strategies to intensify the temporal relationships\textcolor{reb}{, while some other TGNNs only rely on a single strategy. For example, pure memory TGNNs \cite{jodie,dyrep,apan} directly use the node memory as the dynamic node embeddings, potentially with complex $\textup{COMB}$ and $\textup{UPDT}$ function to update node memory. For example, in APAN \cite{apan}, the mails are delivered to the mailboxes of hop-1 neighbors and the $\textup{COMB}$ function applies attention mechanism to update the node memory.}
After studying the architecture of different TGNNs, we identify three components that form a unified representation for most TGNN variants -- the node memory, the attention aggregator, and the temporal sampler. 
\textcolor{reb}{For snapshot-based TGNNs \cite{evolvegcn,dysat}, each snapshot is treated independently while the output of each snapshot is combined to produce the final node embeddings.} Beside these methods, CAW-N~\cite{wang2021inductive} generates high quality dynamic node embeddings by aggregating from multiple causal anonymous walks.

\subsection{Node Memory}

For nodes with different history lengths, a fixed number of temporal neighbors may not provide enough information to generate the dynamic node embedding at the current state. To address this issue, many works \cite{jodie,dyrep,tgn,apan} use node memory to summarize the history of the nodes in the past. Later when this node is referenced as a temporal neighbor, its node memory serves as complimentary information and is combined with the node features as the input node features. 

To maintain the node memory of each node, when an event indicates the appearing of a new edge, a sequence model (RNN or GRU) is used to update the corresponding node memory. 
If there is a new edge connecting from node $u$ to node $v$ 
at the current timestamp $t$, we generate two mails $m_e^{uv}$ and $m_e^{vu}$
\begin{align}
    m_e^{uv} & =\left(\bm s_u||\bm s_v||\Phi(t-t^-_v)||\bm e_{uv}\right)\\
    m_e^{vu} & =\left(\bm s_v||\bm s_u||\Phi(t-t^-_v)||\bm e_{uv}\right).
\end{align}
The time encoder $\Phi$ \cite{tgat} encodes the time interval $\Delta t=t-t^-_v$ into vector
\begin{equation}
    \Phi(\Delta t)=\cos(\bm\omega\Delta t+\bm\phi),
\end{equation}
where $\bm\omega$ and $\bm\phi$ are two learnable vectors. The node memory is then updated by 
\begin{equation}
    \bm s_v=\textup{UPDT}\left(\bm s_v,\textup{COMB}\left(m_e^{ij}|v\in\mathcal{N}(i)\cup\mathcal{N}(j)\right)\right),
    \label{eq:memupdate}
\end{equation}
where 
$\textup{UPDT}$ is the RNN or GRU memory updater, and $\textup{COMB}$ is the combiner of all related neighbor input mails. The mails are delivered to the neighbors of the source and destination nodes. 

When performing GNN message passing, the node memory is combined with the original node features $\bm v_v$ to serve as the new node features.
\begin{equation}
    \bm v'_v=\bm s_v+\textup{MLP}(\bm v_v).
\end{equation}

\subsection{Attention Aggregator}

TGNNs adopt the attention mechanism from Transformer \cite{transformer} to gather and aggregate information from temporal neighbors. The attention aggregation of node $u$ is computed by the queries, keys, and values from its hop-1 temporal neighbors $v\in\mathcal{N}(v)$.

\subsection{Temporal Sampler}

To ensure each node can access the relevant information from its supporting nodes or the mails are delivered to neighbor nodes, TGNNs need to consider the edge timestamps when sampling. There are two major sampling strategies, uniform sampling where neighbors in the past are sampled uniformly as supporting nodes and most-recent sampling where only the most recent neighbors are sampled. Note that in a dynamic graph, two nodes can have multiple edges at different timestamps. These nodes can also be sampled multiple times as supporting nodes with different timestamps.

\section{TGL}

In this section, we present TGL -- a general framework for efficient TGNNs training on large-scale dynamic graphs. 

\label{sec:tgldesign}

\begin{figure}[t]
    \centering
    \input{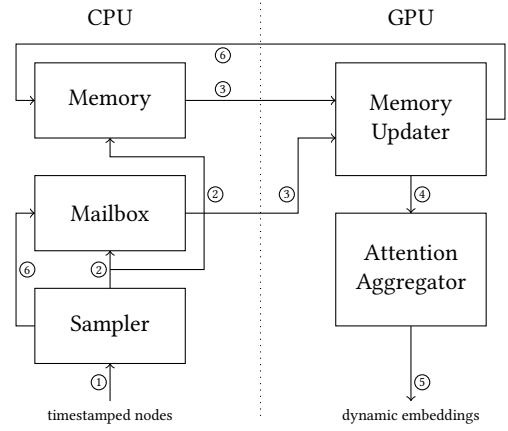}
    \caption{Overview (forward path) of the proposed framework. \textcircled{\small 1} Sample neighbors for the root nodes with timestamps in the current mini-batch. \textcircled{\small 2} Lookup the memory and the mailbox for the supporting nodes. \textcircled{\small 3} Transfer the inputs to GPU and update the memory. \textcircled{\small 4} Perform message passing using the updated memory as input. \textcircled{\small 5} Compute loss with the generated temporal embeddings. \textcircled{\small 6} Update the memory and the mailbox for next mini-batch.}
    \label{fig:tgnn}
\end{figure}

\setlength{\marginparwidth}{15mm}
Figure \ref{fig:tgnn} shows the overview of the training of TGL on a single GPU. We split the modules with learnable and non-learnable parameters to store on GPU and CPU respectively. For datasets where the GPU memory is sufficient to hold all information, the non-learnable modules can also be stored and computed on GPU to speedup the training. 
To be compatible with different TGNN variants, we design five general components: the temporal sampler, the mailbox, the node memory, the memory updater, and the attention aggregator. 
For snapshot-based TGNNs, the temporal sampler would sample in each snapshot individually. Note that in TGL, we do not treat graph snapshots as static windows. Instead, the graph snapshots are dynamically created according to the timestamp of the target nodes. This allows the snapshot-based TGNNs to generate dynamic node embeddings at any timestamps, instead of a constant embedding in a static snapshot.

TGNNs are usually self-supervised by the temporal edges, because it is hard to get dynamic graphs with enough dynamic node labels to supervise the training. Training with temporal edges causes the ``information leak'' problem where the edges to predict are already given to the models as input. The information leak problem in attention aggregator can be simply avoided by not sampling along the edges to predict. In node memory, the information leak problem is eliminated by caching the input from previous mini-batches \cite{tgn}, which enables the node memory to receive gradients. In TGL, we adopt the mailbox module \cite{apan} to store a fixed number of most recent mails for updating the node memory. When a new interaction appears, we first update the node memory of the involved nodes with the cached messages in the mailbox. The messages in the mailbox are updated after the dynamic node embeddings are computed. Note that to keep the node memory consistent, the same updating scheme is used at inference when updating the node memory is not necessary. 

TGL users can easily configure TGL to train different TGNN variants by yaml configuration files. TGL supports a wide range of TGNN variants including vanilla attention-based TGNN TGAT \cite{tgat}, snapshot-based TGNNs like Evolve-GCN \cite{evolvegcn} and DySAT \cite{dysat}, memory-based TGNNs \cite{dyrep,tgn}, and pure memory-based TGNNs \cite{jodie,apan}.

\subsection{Parallel Temporal Sampler}
\label{sec:tglsampler}

\begin{algorithm}[t]
    \SetAlgoLined
    \KwData{sorted T-CSR $G$}
    \KwIn{root nodes $\bm{n}$ with timestamp $\bm{t_n}$, number of layer $L$, number of neighbors in each layer $k_l$, number of snapshots $S$, snapshot length $t_s$}
    \KwOut{DGL MFGs}
    \textcolor{reb}{advance the pointer of $\bm{n}$ to $\bm{t_n}$ in $pt(S+1)$ in parallel\;}
    \For{l \textup{in} 0..L}{
        \For{s \textup{in} 0..S}{
            \If{$l\geq 0$}
            {
              set $\bm{n}$ and $\bm{t_n}$ to sampled neighbors in $l-1$\;
            }
            \eIf{$l==0$}%
            {
                advance the pointer of $\bm{n}$ to $\bm{t_n}-s*t_s$ in $pt(S-s-1)$ in parallel\;
            }{
                binary search in the snapshots $S_s$ for each node $n\in\bm{n}$ in parallel\;
            }
            \ForEach{$n\in\bm{n}$ \textup{in parallel}}{
                sample $k_l$ neighbors within the snapshot $S_s$\;
            }
            generate DGL MFGs\;
        }
    }
    \caption{Parallel Temporal Sampler}
    \label{alg:spl}
\end{algorithm}

Sampling neighbors on dynamic graphs is complex as the \textcolor{reb}{timestamps of the neighbors need to be considered.} 
\textcolor{reb}{In the offline training process, TGL stores the entire dynamic graph statically where the timestamps are attached to the nodes and edges.}
For snapshot-based TGNNs, the temporal samplers need to identify the snapshots before sampling. \textcolor{reb}{Other TGNNs that either samples uniformly from all past neighbors or sample most recent neighbors can be treated as single snapshot TGNNs with infinite snapshot length. Their temporal samplers also needs to identify the candidate edges and their sampling probabilities. Hence, it is important to design a data structure that can rapidly identifies the dynamic candidate set of temporal neighbors to sample from.} Combined with the fact that the mini-batches in TGNNs training follow chronological order (have non-decreasing timestamps), we propose the Temporal-CSR (T-CSR) data structure. 

\begin{figure}[t]
    \centering
    \input{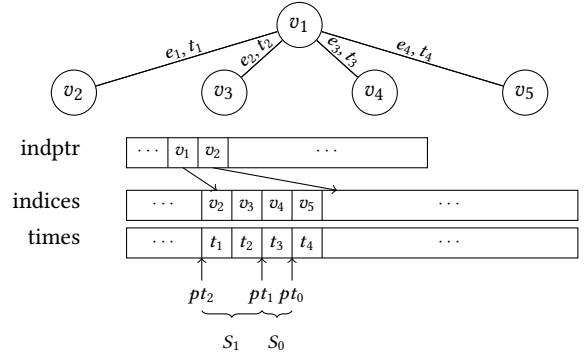}
    \caption{T-CSR representation of the node $v_1$ with four temporal edges $e_1$ to $e_4$ with timestamps $t_1$ \textcolor{reb}{to} $t_4$ connected to neighbors $v_1$ to $v_4$. \textcolor{reb}{The indices and times arrays are sorted by the edge timestamps and indexed by the edge ids $e_1$ to $e_4$.} $S_0$ and $S_1$ denote two snapshots of the temporal graph, designated by the pointers $pt_0$ to $pt_2$.}
    \label{fig:dssample}
\end{figure}

\noindent \textcolor{reb}{\textbf{The T-CSR Data Structure} Besides the indptr and indices array of the CSR data structure, for each node, T-CSR sorts the outgoing edges according to their timestamps as shown in Figure \ref{fig:dssample}. After sorting all the edges in a dynamic graph, we assign edge ids according to their position (indexes) in the sorted indices and times arrays.}
\textcolor{reb}{In addition, for a TGNN model with $n$ snapshots, we maintain $n+1$ pointers for each node that point at the first and last edges in these snapshots. Formally, the T-CSR data structure is defined by an indptr array of size $|V|+1$, an indices array and a time array of size $|E|$, and $n+1$ pointers array of size $|V|$, which leads to a total space complexity of $\mathcal{O}(2|E|+(n+2)|V|)$.}
\textcolor{reb}{For dynamic graphs with inserting, updating, and deletion of edges and nodes, the T-CSR data structure can treat them as standalone graph events and allocate their own entries in the indices and times array.}

\noindent \textcolor{reb}{\textbf{Sampling} With the help of the T-CSR data structure, we can quickly choose an edge between two pointers uniformly, or pick edges closest to the end pointer for the most recent neighbors. }
These pointers are stored in arrays and takes an additional $\mathcal{O}(n|V|)$ storage and $\mathcal{O}(|E|)$ computation complexity to maintain in one epoch, but allows the sampler to identify candidate edges in $\mathcal{O}(1)$. By contrast, performing binary search would lead to $\mathcal{O}(|E|\log|E|)$ computation complexity to identify candidate edges in one epoch. 
Note that some TGNNs like TGAT \cite{tgat} use the timestamp of the neighbors to sample multi-hop temporal neighbors, instead of using the timestamp of the root nodes. For these TGNNs, the proposed pointer only works for the hop-1 neighbors. Since the edges are sorted in T-CSR, we can still to use binary search to quickly find out the candidate edges before sampling.

\noindent \textcolor{reb}{\textbf{Parallel Sampling}} To leverage the multi-core CPU resources in the host machine, \textcolor{reb}{we exploit data parallelism to sample on the root nodes in a mini-batch as shown in Algorithm \ref{alg:spl}.}
In each mini-batch, \textcolor{reb}{the target nodes are evenly distributed to each thread to update the pointers and sample the neighbors. Note that when updating the pointers in parallel, it is possible that multiple threads share the same target nodes with different timestamps, which causes race conditions. We add fine-grained locks to each node to avoid the pointers being advanced multiple times under such conditions.}
When same target nodes at different timestamp appears multiple time in one mini-batch, it is also possible that the target nodes with small timestamps sample temporal neighbors from the future. We prevent information leak in such situation by strictly enforcing that the sample temporal neighbors have smaller timestamps than the root nodes.
\textcolor{reb}{After each thread finishes sampling in each mini-batch, we generate a DGL Message Flow Graph (MFG) for each layer \cite{wang2019dgl} which contains all the input data needed in the forward and backward propagation and pass it to the trainer.}

\subsection{\textcolor{reb}{Parallel Training}}
\label{sec:tglngpu}

\begin{algorithm}[t]
    \textcolor{reb}
        {
            \SetAlgoLined
            \KwData{training edges $E$, sorted T-CSR $G$, TGNN model $M$}
            \KwIn{batch size $bs$, chunk size $cs$, training epochs $E$}
            \For{e \textup{in} 0..E}{
                $e_s\leftarrow\mbox{rand}(0,bs/cs)*bs$\;
                $e_e\leftarrow e_s+bs$\;
                \While{$e_e\leq|E|$}{
                    \textup{sample MFGs from }$E(e_s..e_e)$\;
                    \textup{train for one iteration with the current MFG}\;
                    $e_s\leftarrow e_s+bs$\;
                    $e_e\leftarrow e_e+bs$\;
                }
            }
            \caption{Random Chunk Scheduling}
        }
    \label{alg:shce}
\end{algorithm}

\textcolor{reb}{In order to scale static GNN training to large graphs, recent works \cite{distdgl,flexgraph} increase the batch size to take advantage of the massive data parallelism provided by multi-GPU servers or GPU clusters. However, training TGNN with a large batch size suffers from the intrinsic temporal dependency in the node memory. Defining the dependent edges as pairs of training edges who share common supporting nodes in the source or destination nodes, we can divide the edge dependencies into two types:}
\begin{itemize}
    \item \textbf{Intra-batch} dependencies refer to the dependent edges in the same mini-batch. In TGNN training, the intra-batch dependencies are discarded in order to process the edges in a mini-batch in parallel.
    \item \textbf{Inter-batch} dependencies refer to the dependent edges in different mini-batches. TGNNs take these inter-batch relations into account by updating the node memory and the mailbox after each mini-batch.
\end{itemize}
\textcolor{reb}{Since the total number of intra- and inter-batch dependencies is constant on one dynamic graph, training with a larger batch size discards more intra-batch dependencies and learns less inter-batch dependencies which leads to lower accuracy.
To mitigate this issue, we propose a random chunk scheduling technique that divides the training edges into chunks and randomly picks one chunk as the starting point in each training epoch, which allows close chunks to be arranged in different mini-batches in different training epochs, hence learning more inter-batch dependencies. The random chunk training algorithm is shown in Algorithm \ref{alg:shce}.}

\textcolor{reb}{To train TGL on multiple GPUs, we adopt the setup of multiple GPUs on a single node and store the node memory and the mailbox in the main memory. On $n$ GPUs, we launch $n$ training processes and one sampling process with inter-process communication protocols. For simplicity, the updates of the model weights, the node memory, and the mailbox are synchronized.}

\section{Experiments}

We perform detailed experiments to evaluate the performance of TGL. 
We implement TGL using PyTorch 1.8.1 \cite{pytorch} and DGL 0.6.1 \cite{wang2019dgl}. The parallel temporal sampler is impletmented using C++ and integrated to the Python training script using PyBind11 \cite{pybind11}. The open-sourced code of TGL could be found at \url{https://github.com/tedzhouhk/TGL}.

We select five representative TGNN variants as the baseline methods and evaluate their performance in TGL. 
\begin{itemize}
    \item JODIE \cite{jodie} is a pure memory-based TGNN method that uses RNN to update the node memory by the node messages. We use the open-sourced code implemented as a baseline in TGN \cite{tgn} as the baseline code.
    \item DySAT \cite{dysat} is a snapshot-based TGNN that uses RNN to combine the node embeddings from different snapshots.
    \item TGAT \cite{tgat} is a attention-based TGNN that gathers temporal information by the attention aggregator.
    \item TGN \cite{tgn} is a memory-based TGNN that applies the attention aggregator on the node memory updated by GRU with the node messages.
    \item APAN \cite{apan} is a pure memory-based TGNN method that uses attention aggregator to update the node memory by the node messages delivered to the multi-hop neighbors.
\end{itemize}
For fair comparison, we set the receptive field to be 2-hop and fix the number of neighbors to sampler per hop at 10. The size of the mailbox is set to be 10 mails in APAN while 1 mail in other methods. %
For the COMB function in Equation \ref{eq:memupdate}, we use the most recent mail in all methods, as we do not see noticeable difference if switched to the mean of mails. We set the dimension of the output dynamic node embeddings to be 100. We apply the attention aggregator with 2 attention heads for the message passing step in all baseline methods. For DySAT, we use 3 snapshots with \textcolor{reb}{the duration of each snapshot to be 10000 seconds on the four small-scale datasets, 6 hours on GDELT, and 5 years on MAG.} As mentioned in Section \ref{sec:tgldesign}, TGL uses dynamic snapshot windows to ensure that the time resolution of the generated dynamic node embeddings is the same as the other TGNNs. \textcolor{reb}{For fairness, we add layer normalization to JOIDE and TGAT, which allows all methods to have layer normalization and in-between each layer.} For all methods, we sweep the learning rate from \{0.01,0.001,0.0001\} and dropout from \{0.1,0.2,0.3,0.4,0.5\}. The TGNN models are trained with the link prediction task and directly used in the dynamic node classification task without fine-tuning \cite{tgat,tgn}. On all datasets, we follow the extrapolation setting that predict the links or node properties in the future given the dynamic graphs in the past. We provide comprehensive and nondiscriminatory benchmark results for various TGNNs by evaluating them in the TGL framework.

\subsection{Datasets}
\label{sec:expds}

\begin{table}[t]
  \setlength{\tabcolsep}{0.9mm}
  \centering
  \caption{Dataset Statistic. The max$(t)$ column shows the maximum edge timestamp (minimum edge timestamp is 0 in all datasets). $|d_v|$ and $|d_e|$ show the dimensions of node features and edge features, respectively. \textcolor{reb}{The * denotes randomized features.}}
  \begin{tabular}{r|ccccccc}
        & $|V|$ & $|E|$ & max$(t)$ & Labels & Classes & $|d_v|$ & $|d_e|$\\
      \toprule
      Wikipedia & 9K & 157K & 2.7e6 & 217 & 2 & - & 172\\
      Reddit & 11K & 672K & 2.7e6 & 366 & 2 & - & 172\\
      \textcolor{reb}{MOOC} & \textcolor{reb}{7K} & \textcolor{reb}{412K} & \textcolor{reb}{2.6e6} & \textcolor{reb}{-} & \textcolor{reb}{-} & \textcolor{reb}{-} & \textcolor{reb}{128*}\\
      \textcolor{reb}{LastFM} & \textcolor{reb}{2K} & \textcolor{reb}{1.3M} & \textcolor{reb}{1.3e8} & \textcolor{reb}{-} & \textcolor{reb}{-} & \textcolor{reb}{-} & \textcolor{reb}{128*}\\
      \midrule
      GDELT & 17K & 191M & 1.8e5 & 42M & 81 & 413 & 186\\
      MAG & 122M & 1.3B & 120 & 1.4M & 152 & 768 & -\\
  \end{tabular}
  \label{tab:ds}
\end{table}

Table \ref{tab:ds} shows the statistic of the six datasets we use to evaluate the performance of TGL. \textcolor{reb}{As the Wikipedia \cite{tgn}, Reddit \cite{tgn}, MOOC \cite{jodie}, and LastFM \cite{jodie} datasets are small-scale and bipartite dynamic graphs, in order to evaluate the performance on general and large-scale graphs, we introduce} two large-scale datasets -- GDELT and MAG. These two datasets contains 0.2 and 1.3 billion edges in multiple years and focus on testing the capability of TGNNs in two different dimensions.

\subsubsection{GDELT}

The GDELT dataset is a Temporal Knowledge Graph (TKG) originated from the Event Database in GDELT 2.0 \cite{gdelt} which records events happening in the world from news and articles in over 100 languages every 15 minutes. Previous event prediction work \cite{renet} pre-processed a small dataset from the same source with events happened in January 2018. Their version is a featureless graph where the features of actors and events are ignored. In this work, we propose a larger featured version with events happened from the beginning of 2016 to the end of 2020. Our GDELT dataset is a homogeneous dynamic graph where the nodes represent actors and temporal edges represent point-time events. Each node has a 413-dimensional multi-hot vector representing the CAMEO codes attached to the corresponding actor to server as node features. Each temporal edge has a timestamp and a 186-dimensional multi-hot vector representing the CAMEO codes attached to the corresponding event to server as temporal edge features. The link prediction task on the GDELT dataset predicts whether there will be an event happening between two actors at a given timestamp. For the node classification task, we use the countries where the actors were located when the events happened as the dynamic node labels. We remove the dynamic node labels for the nodes that have the same labels at their most recent timestamps to make this task more challenging. We use the events before 2019, in 2019, and in 2020 as training, validation, and test set, respectively. 
The GDELT datasets has dense temporal interactions between the nodes and requires TGNNs to be able to capture mutable node information for a long time duration.

\subsubsection{MAG}

The MAG dataset is a homogeneous sub-graph of the heterogeneous MAG240M graph in OGB-LSC \cite{ogblsc}. We extract the paper-paper citation network where each node in MAG represents one academic paper. A directional temporal edge from node $u$ to node $v$ represents a citation of the paper $v$ in the paper $u$ and has a timestamp representing the year when the paper $u$ is published. The node features are 768-dimensional vectors generated by embedding the abstract of the paper using RoBERTa \cite{liu2019roberta}. The link prediction task on the MAG dataset predicts what papers will a new paper cite. For the node classification dataset, we use the arXiv subject areas as node labels. We use the papers published before 2018, in 2018, and in 2019 as training, validation, and test set.
The MAG dataset test the capability of TGNN models to learn dynamic node embeddings on large graph with stable nodes and edges.

\begin{table}[t]
  \setlength{\tabcolsep}{0.75mm}
  \centering
  \caption{Execution time and improvement with respect to baseline samplers on the Wikipedia dataset for one epoch.}
  \begin{tabular}{r|ccc|ccc|ccc}
     & \multicolumn{3}{c|}{DySAT} & \multicolumn{3}{c|}{TGAT} & \multicolumn{3}{c}{TGN}\\
    \hline
    \#Threads & 1 & 8 & 32 & 1 & 8 & 32 & 1 & 8 & 32\\
    \toprule
    Time (s) & 1.161 & 0.446 & 0.371 & 1.557 & 0.569 & 0.370 & 0.094 & 0.46 & 0.039\\
    Improv. & - & - & - & 23$\times$ & 48$\times$ & 57$\times$ & 69$\times$ & 188$\times$ & 289$\times$\\
  \end{tabular}
  \label{tab:samp}
\end{table}

\begin{table*}[ht!]
  \centering
  \caption{Link Prediction results on \textcolor{reb}{the Wikipedia, Reddit, MOOC, and LastFM datasets. } The Time columns refer to the training time per epoch. (\textbf{First} \textnormal{\underline{second})}}
  \setlength{\tabcolsep}{3pt}
  \begin{tabular}{r|cc|ccc|cc|ccc||cc|cc}
     & \multicolumn{5}{c|}{Wikipedia} & \multicolumn{5}{c||}{Reddit} & \multicolumn{2}{c|}{\textcolor{reb}{MOOC}} & \multicolumn{2}{c}{\textcolor{reb}{LastFM}}\\
     & \multicolumn{2}{c|}{Baseline} & \multicolumn{3}{c|}{TGL} & \multicolumn{2}{c|}{Baseline} & \multicolumn{3}{c||}{TGL} & \multicolumn{2}{c|}{\textcolor{reb}{TGL}} & \multicolumn{2}{c}{\textcolor{reb}{TGL}}\\
     & AP & Time (s) & AP & Time (s) & Speedup & AP & Time (s) & AP & Time (s) & Speedup & \textcolor{reb}{AP} & \textcolor{reb}{Time (s)} & \textcolor{reb}{AP} & \textcolor{reb}{Time (s)}\\
     \toprule
    JODIE & 94.35 & 16.6 & \underline{98.90} & \textbf{1.0} & 16.94$\times$ & 96.56 & 89.0 & 99.45 & \textbf{4.2} & 21.24$\times$ & \textcolor{reb}{\underline{98.95}} & \textcolor{reb}{\textbf{2.8}} & \textcolor{reb}{\textbf{78.78}} & \textcolor{reb}{\textbf{8.7}}\\
    DySAT & - & - & 96.37 & 6.4 & - & - & - & 98.57 & 21.5 & - & \textcolor{reb}{98.76} & \textcolor{reb}{19.5} & \textcolor{reb}{\underline{76.39}} & \textcolor{reb}{48.4}\\
    TGAT & 95.09 & 110.1 & 97.26 & 6.6 & 16.73$\times$ & 97.82 & 576.2 & \underline{99.48} & 39.9 & 14.45$\times$ & \textcolor{reb}{98.50} & \textcolor{reb}{24.5} & \textcolor{reb}{54.82} & \textcolor{reb}{91.4}\\
    TGN & 98.34 & 17.7 & \textbf{99.62} & 2.1 & 8.51$\times$ & 98.47 & 91.9 & \textbf{99.78} & 10.5 & 8.33$\times$ & \textcolor{reb}{\textbf{99.59}} & \textcolor{reb}{5.7} & \textcolor{reb}{73.76} & \textcolor{reb}{18.7}\\
    APAN & 98.12 & 8.8 & 98.14 & \underline{2.0} & 4.38$\times$ & 99.22 & 121.7 & 99.24 & \underline{8.8} & 13.85$\times$ & \textcolor{reb}{98.58} & \textcolor{reb}{\underline{5.6}} & \textcolor{reb}{62.73} & \textcolor{reb}{\underline{18.2}}\\
  \end{tabular}
  \label{tab:1gpu}
\end{table*}

\subsection{Parallel Temporal Sampler}
\label{sec:expsample}

The performance of our parallel temporal sampler is evaluated on the g4dn.8xlarge instance on AWS EC2 with 32 virtual CPUs and 64GB of main memory. We select three representative sampling 
algorithms  
\begin{itemize}
    \item DySAT 2-layer sampling represents the temporal graph sampling for snapshot-based methods. The supporting nodes are chosen uniformly from the temporal neighbors in each dynamic snapshots.
    \item TGAT 2-layer sampling represents the uniformly temporal graph sampling which selects supporting nodes uniformly from all past temporal neighbors.
    \item TGN 1-layer sampling represents the most recent temporal graph sampling which selects most recent temporal neighbors as supporting nodes. Most recent sampling algorithms are usually used in memory-based methods and hence requires one less supporting layers.
\end{itemize}

\begin{figure}[b]
  \centering
  \input{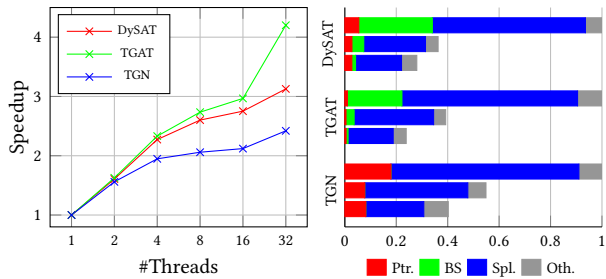}
  \caption{(a) Scalability of the temporal sampler on the Wikipedia dataset. (b) Runtime breakdown (normalized by single thread runtime) of the temporal sampler on the Wikipedia dataset with 1 (top), 8 (mid), and 32 (bottom) threads. Ptr., BS, Spl., and Oth. denote the time to update pointers (line 4 and 11), to perform binary search (line 13), to sample neighbors (line 16), and to generate DGL MFGs (line 18) in Algorithm \ref{alg:spl}, respectively.}
  \label{fig:samp}
\end{figure}

Table \ref{tab:samp} shows the improvement (speedup) of the temporal parallel sampler in TGL compared with the samplers in the open-sourced baselines using different number of threads. \textcolor{reb}{The baseline samplers sample the neighbors by performing single-thread vectorized binary search on sorted neighbors lists. We show the sampling time for one epoch with batch size of 600 positive and 600 negative edges.} With our efficient T-CSR data structure, TGL \textcolor{reb}{spends} less than 0.5 seconds in sampling on one epoch of the Wikipedia dataset for all three sampling algorithms. Using 32 threads, TGL achieves $57\times$ and $289\times$ speedup compared with the sampler in TGAT and TGN. 
\textcolor{reb}{The speedup is a result by combined factors of 1) the T-CSR data structure, 2) data parallelism, and 3) efficiency of C++ over Python.}

Figure \ref{fig:samp} shows the runtime and the runtime breakdown of our temporal parallel sampler using a different number of threads. TGL achieves $3.13\times$, $4.20\times$, and $2.42\times$ speedup using 32 threads for the DySAT, TGAT, and TGN sampling algorithms. The reasons for the sub-linear speedup are 1) node-wise locks in updating the pointers 2) memory performance bottleneck when fetching the selected edge information 3) linear workload with respect to the number of threads when generating DGL MFGs.

\subsection{Single-GPU Training}
\label{sec:expsmall}

We evaluate the performance of TGL with a single GPU on the \textcolor{reb}{four} small datasets. We use the same g4dn.8xlarge AWS EC2 instance with one Nvidia T4 GPU. We find that on \textcolor{reb}{all datasets}, the batch size of 600 positive edges with 600 negative edges are a good balance point between the convergence rate and training speed for memory-based TGNNs. Hence, \textcolor{reb}{for a fair comparison}, we use batch size of 600 for all five selected TGNN variants in TGL and their open-sourced baselines. \textcolor{reb}{For the MOOC and LastFM datasets, we randomly generate 128-dimensional edge features since the original datasets do not contain node or edge features. Due to the lack of authentic features and performance evaluation of the baseline code, we do not compare the performance of TGL and the baseline code on these two datasets.} Since the 16GB GPU memory is enough to hold the node features, the edge features, the node memory, and the mailbox, we store these data on GPU instead of CPU to avoid the data transfer overhead. We use 32 threads in the temporal parallel sampler.

\begin{figure}[b]
    \centering
    \input{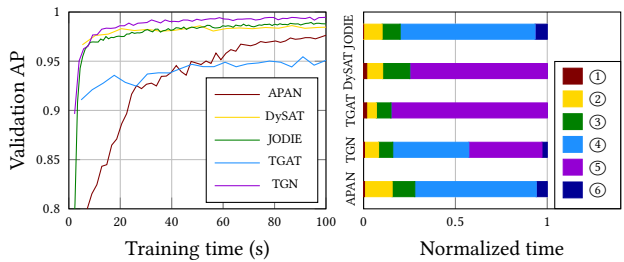}
    \caption{Validation AP with training time (left) and \textcolor{reb}{normalized} runtime breakdown (right) on the Wikipedia dataset. The circled numbers
    refer to the \textcolor{reb}{six} steps in Figure \ref{fig:tgnn}.}
    \label{fig:wiki}
\end{figure}

Table \ref{tab:1gpu} shows the accuracy and per epoch training time of the five baselines and TGL in the link prediction task. We report the accuracy in Average Precision (AP) on both the positive and negative test edges. For all methods, TGL achieves similar or higher AP than the baselines with significantly faster runtime (see Figure \ref{fig:aptime}). 
\textcolor{reb}{The accuracy improvement on TGAT and JODIE is because we use layer normalization in-between each layer. The accuracy of TGAT and TGN also benefits from better hyper-parameters and convergence. TGN achieves the highest AP in the link prediction task on all datasets except the LastFM dataset, followed by JODIE, DySAT and TGAT. The pure memory-based TGNN and JODIE achieves top-tier accuracy with the fastest training time.} 
With efficiently implemented components and optimized data path, TGL achieves an average of $13\times$ speedup in the per-epoch training time.

Figure \ref{fig:wiki} shows the convergence curve and runtime breakdown on the Wikipedia dataset. JODIE, DySAT, and TGN have faster convergence speed than APAN and TGAT. \textcolor{reb}{The runtime breakdown is measured by storing the node memory and mailbox in the main memory and forcing to use synchronized execution between the CPU and GPU, which leads to around 15\% more execution time than asynchornized execution.} With our temporal parallel sampler, the sampling overhead in TGL is negligible. For computation intensive two-layer TGNNs like DySAT and TGAT, the runtime is dominated by the computation on GPU. For memory-based models, the time spent in updating the node memory and the mailbox takes up to 30\% of the total training time.

Table \ref{tab:nodeclass} shows the results of directly using the learned TGNN models on the dynamic node classification task. On the Wikipedia and the Reddit datasets, the node classification tasks are to identify banned users. Since the number of positive labels are small compared with the number of negative labels, we train the MLP classifiers with an equal number of randomly sampled negative labels, similar to training link prediction models. We also show the accuracy as AP on both the positive nodes and sampled the negative nodes. TGN and JODIE achieve the highest AP on the Wikipedia and the Reddit datasets, where JODIE achieves more than 7\% AP than other methods on the Reddit dataset. We assume this is due to the noisy neighbors in the Reddit dataset, which prevent high-expressive model from learning general patterns on the graph structure.

\begin{table}[b]
  \centering
  \caption{Dynamic node classification result (\textbf{First} \textnormal{\underline{second})}).}
  \begin{tabular}{r|cc|cc}
     & Wikipedia & Reddit & GDLET & MAG \\
    \hline
     & \multicolumn{2}{c|}{AP} & \multicolumn{2}{c}{F1-Micro}\\
    \toprule
    JODIE & 81.73 & \textbf{70.91} & \underline{11.25} & 43.94\\
    DySAT & \underline{86.30} & 61.70 & 10.05 & \underline{50.42}\\
    TGAT & 85.18 & 60.61 & 10.04 & \textbf{51.72}\\
    TGN & \textbf{88.33} & \underline{63.78} & \textbf{11.89} & 49.20\\
    APAN & 82.54 & 62.00 & 10.03 & -\\
  \end{tabular}
  \label{tab:nodeclass}
\end{table}

\subsection{Random Chunk Scheduling}
\label{sec:exprand}

\begin{table}[b]
  \centering
  \caption{Link Prediction results of TGL on GDELT and MAG. The Time columns refer to the training time per epoch (\textbf{First} \textnormal{\underline{second})}).}
  \begin{tabular}{r|cc|cc}
     & \multicolumn{2}{c|}{GDELT} & \multicolumn{2}{c}{MAG}\\
    \hline
     & AP & Time (s) & AP & Time (s) \\
    \toprule
    JODIE & 97.98 & \textbf{599.2} & \underline{99.41} & \textbf{4128.3}\\
    DySAT & \underline{98.72} & 10651.4 & 98.27 & 19748.6\\
    TGAT & 96.49 & 8499.2 & 99.02 & 32104.5\\
    TGN & \textbf{99.39} & \underline{915.9} & \textbf{99.49} & \underline{8912.5}\\
    APAN & 95.28 & 1358.5 & - & -\\
  \end{tabular}
  \label{tab:8gpus}
\end{table}

\begin{figure}[t]
  \centering
  \input{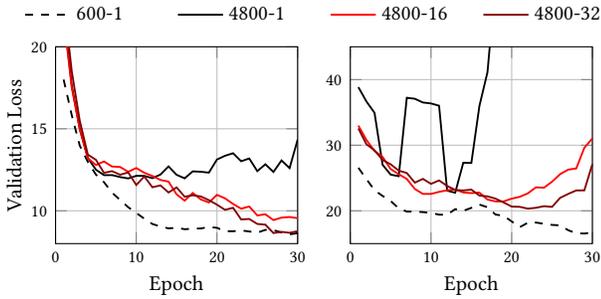}
  \caption{Validation loss (moving average of 5 epochs) with different chunk size when using random chunk scheduling algorithm with large batch size on Wikipedia (left) and Reddit (right) dataset. We denote batch size $x$ and number of chunks per batch $y$ as $x-y$ in the legends.}
  \label{fig:batchsize}
\end{figure}

To evaluate the effectiveness of the random chunk scheduling technique, we train the TGN model which has the best overall performance on the two small-scale datasets, as training with a small batch size and plot various convergence curves on the large-scale datasets is too slow. To make a fair comparison, we train the baseline models with the best group of hyperparameters (0.001 learning rate, 600 batch size). We then increase the batch size and also linearly increase the learning rate, as a larger batch size leads to a better approximation of the total loss \cite{goyal2017accurate}. Specifically, we train the same model with 8$\times$ the batch size and learning rate (0.008 learning rate, 4800 batch size) with the chunk sizes of 4800, 300, and 150 (number of chunks per batch size 1, 16, and 32). Since the node memory used in the validation process is inherited from the training process, when we compute the validation loss, we first reset the node memory and use a constant batch size of 600 to run one whole epoch on the training and validation set. Figure \ref{fig:batchsize} shows the validation loss under different batch sizes and chunk sizes on the Reddit and Wikipedia datasets. The models trained with the batch size of 4800 and no random chunk scheduling cannot learn on both datasets after 5 to 10 epochs, due to the lost dependencies in the training mini-batches. On the Wikipedia dataset, the batch size of 4800 with 16 chunks per batch performs better than no chunks while the same batch size with 32 chunks per batch achieves similar convergence after 30 epochs. On the Reddit dataset, our random chunk scheduling technique also mitigates the overfitting issue and achieves close to baseline convergences within 30 epochs.

\subsection{Multi-GPU Training}
\label{sec:explarge}

We evaluate the performance of TGL on the two large-scale datasets with multiple GPUs. We use the p3dn.24xlarge instance on EC2 with 96 virtual CPUs, 768GB of main memory, and 8 Nvidia V100 GPUs. We use 64 threads in the temporal parallel sampler and assign 8 threads for each trainer process. We use a local batch size of 4000 positive and 4000 negative edges on each GPU. The global copy of the node memory and the mailbox are stored in the shared memory. %
The trainer process then overlaps the MFG copy to GPU with the computation on GPU by creating additional copying threads \textcolor{reb}{on} different CUDA streams. The gradients in each iteration are synchronized among the trainer processes through the NCCL backend. 

\begin{figure}[t]
    \centering
    \input{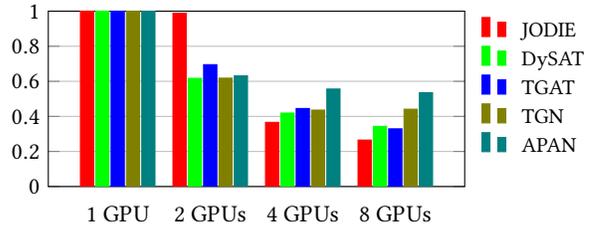}
    \caption{Normalized Training time per epoch with different number of GPUs on the GDELT dataset.}
    \label{fig:gpuscale}
\end{figure}

Table \ref{tab:8gpus} shows the AP and running time in the link prediction task. Similar to the single GPU results, TGN achieves the highest AP and JODIE has the fastest training time. On the GDELT datasets, the memory-based models can train one epoch within 30 minutes, while the non-memory based models need more than 3 hours. 
On the MAG dataset, APAN throws out of memory error as it requires the mailbox to store 10 most recent mails for each node in the graph. Figure \ref{fig:gpuscale} shows the scalability of TGL on multiple GPUs. TGL achieves $2.74\times$, $2.28\times$, $2.25\times$, $2.30\times$ and $1.80\times$ speedup by using 4 GPUs for JODIE, DySAT, TGAT, TGN, and APAN, respectively. For 8 GPUs, the bandwidth between CPU and main memory to slice the node and edge features and update the node memory and the mailbox and the number of PCI-E channels 
to copy the MFGs to the GPUs are saturated. 

Table \ref{tab:nodeclass} shows the F1-Micro of the trained models in the multiple-class single-label dynamic node classification task. On the GDELT dataset, all models perform bad where JODIE and TGN has slightly better performance than others. On the MAG dataset, TGAT and DySAT with two complete graph attention layers achieves the highest and second highest accuracy while JODIE with no graph attention layer achieves the lowest accuracy. 
\section{Conclusion}

In this work, we proposed TGL -- the first unified framework for \textcolor{reb}{large-scale TGNN training. TGL allows users to efficiently train different TGNN variants on a single GPU and multiple GPUs by writing simple configuration files. We designed the T-CSR data structure to store the dynamic graphs and developed a temporal parallel sampler which greatly reduces the sampling overhead.} We proposed the random chunk scheduling technique to mitigate the loss of dependencies when training with a large batch size. We processed two large-scale datasets to test the capability of TGNNs in two different dimensions. We evaluated the performance of five different TGNN variants on
\textcolor{reb}{four small-scale datasets and two large-scale datasets with billions of edges. TGL achieves similar or better accuracy on all datasets with significantly faster training time compared with the open-sourced baselines.}

\bibliographystyle{ACM-Reference-Format}
\bibliography{sample}

\end{document}